\pdfoutput=1

\documentclass[11pt]{article}

\usepackage[]{ACL2023}

\usepackage{times}
\usepackage{latexsym}
\usepackage{arydshln} 

\usepackage[T1]{fontenc}

\usepackage[utf8]{inputenc}

\usepackage{microtype}

\usepackage{inconsolata}
\usepackage{booktabs}
\usepackage{graphicx}
\usepackage{soul}
\usepackage[dvipsnames]{xcolor}
\usepackage[most]{tcolorbox}
\usepackage{tabularx}
\usepackage{enumitem}
\usepackage{float}
\usepackage{bbding}
\usepackage{pifont}

\title{\textsc{SagaQA}: A Multi-hop Reasoning Benchmark for Long-form \\Narrative Understanding in TV Series}

\author{
\textbf{
Galann Pennec\textsuperscript{$\infty,\diamondsuit,\heartsuit$} 
\ \quad
Zhengyuan Liu\textsuperscript{$\diamondsuit$} 
\ \quad 
}
\\
\textbf{
Nicholas Asher\textsuperscript{$\S$} 
\ \quad 
Philippe Muller\textsuperscript{$\infty$} 
\ \quad 
Nancy F. Chen\textsuperscript{$\diamondsuit$}
}
\\
\textsuperscript{$\infty$}IRIT, University of Toulouse, France\\
\textsuperscript{$\diamondsuit$}Agency for Science, Technology and Research (A*STAR), Singapore\\
\textsuperscript{$\heartsuit$}CNRS@CREATE, Singapore
\quad
\textsuperscript{$\S$}CNRS, IRIT, France\\
\texttt{galann.pennec@cnrsatcreate.sg},
\ \texttt{\{liu\_zhengyuan,nancy\_chen\}@a-star.edu.sg}\\
\ \texttt{\{nicholas.asher,philippe.muller\}@irit.fr}
}

\begin{document}

\newcommand{\cit}{{\color{red} cite }}

\definecolor{yellow_fig}{RGB}{255,242,204}
\definecolor{green_fig}{RGB}{226,240,217}
\definecolor{blue_fig}{RGB}{218,227,243}

\newcommand{\hlred}[1]{{\sethlcolor{red}\hl{#1}}}
\newcommand{\hly}[1]{{\sethlcolor{yellow}\hl{#1}}}
\newcommand{\hlgr}[1]{{\sethlcolor{green}\hl{#1}}}
\newcommand{\boxred}[1]{{\fcolorbox{red}{white}{#1}}}

\newcommand{\hlyfig}[1]{{\sethlcolor{yellow_fig}\hl{#1}}}
\newcommand{\hlgrfig}[1]{{\sethlcolor{green_fig}\hl{#1}}}
\newcommand{\hlblue}[1]{{\sethlcolor{blue_fig}\hl{#1}}}

\newcommand{\cmark}{\textcolor{green!70!black}{\ding{52}}} 
\newcommand{\xmark}{\textcolor{red!80!black}{\ding{56}}}   
\newcommand{\mymixed}{\textcolor{orange}{$\boldsymbol{\approx}$}}
\newcommand{\greenn}[1]{\textcolor{green!70!black}{\textbf{#1}}}
\newcommand{\redd}[1]{\textcolor{red!80!black}{\textbf{#1}}}
\newcommand{\orangee}[1]{\textcolor{orange}{\textbf{#1}}}

\newcommand{\coluline}[2]{%
  {\setulcolor{#1}\ul{#2}}%
}

\maketitle
\begin{abstract}

We introduce~\textsc{SagaQA}, a long-form video benchmark for multi-hop reasoning over full-length TV series.
Existing video reasoning benchmarks often emphasize local understanding of adjacent frames or clips.~\textsc{SagaQA} addresses this gap by requiring high-level comprehension of extended multimodal narratives in entire TV shows.
A distinguishing feature of~\textsc{SagaQA} is the granularity of its reasoning steps. Our dataset necessitates long-range reasoning hops to connect information across completely different episodes. This requires models to reason over entire events and actions, demanding a deep understanding of the show’s narration and progression at a multimodal level.
Motivated by recent progress in agentic methods, we further study how different planning strategies handle such complex reasoning. We categorize these approaches into three classes—\emph{Parallel}, \emph{Sequential}, and \emph{Hybrid} planners—and evaluate their ability to generate coherent and complete reasoning plans. Our results on~\textsc{SagaQA} suggest that hybrid planners consistently produce higher-quality plans and exhibit stronger capabilities for complex, high-level narrative understanding in TV shows.\footnote{Our code and benchmark are available at~\url{https://github.com/galannp/sagaqa}}

\end{abstract}

\begin{figure*}[hbt]
\begin{center}
\includegraphics[width=0.88\textwidth]{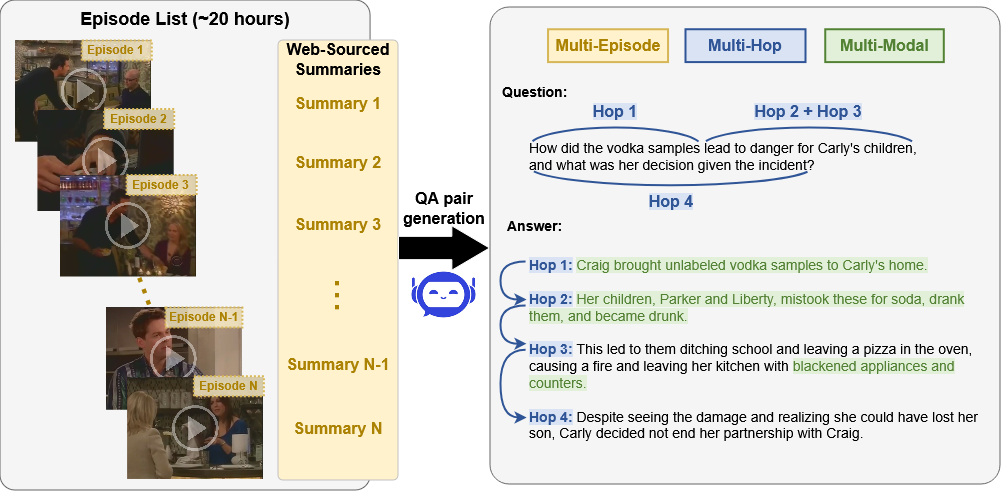}
\end{center}
\caption{\textbf{Our data generation pipeline for~\textsc{SagaQA}.} 1) Given a list of 20 consecutive episodes~$\mathcal{E}_{20}$ and their corresponding annotated summaries from SummScreen3D, we prompt an LLM to generate synthetic QA pairs that satisfy three key criteria:~\hlyfig{\emph{multi-episode}},~\hlblue{\emph{multi-hop}} and~\hlgrfig{\emph{multi-modal}}. In the example shown, the generated question requires four reasoning hops, each being logically connected to the previous one in the reasoning chain. The hops are annotated in \hlblue{BLUE} while the part of the answer that refers to both modalities (video and dialogue) is highlighted in \hlgrfig{GREEN}.}
\label{fig:pipeline}
\vspace{-0.3cm}
\end{figure*}

\section{Introduction}

Recent advancements in Multimodal Large Language Models (MLLMs) have significantly expanded the frontiers of video understanding, with models demonstrating impressive capabilities in visual reasoning~\citep{DBLP:journals/corr/abs-2503-21776, DBLP:journals/corr/abs-2508-18265, DBLP:journals/corr/abs-2507-07966}. However, a critical disparity remains between the ability to perform short-term visual perception and the capacity for long-term narrative comprehension. While existing benchmarks have successfully formalized stepwise causal reasoning within short videos of a few minutes~\citep{DBLP:journals/corr/abs-2505-21374, DBLP:journals/corr/abs-2507-16878} or multi-step reasoning over longer single-file narratives~\citep{DBLP:journals/corr/abs-2506-10857}, they typically rely on adjacent clips and low-level understanding.

To bridge the gap between short-context perception and long-form understanding, recent methodologies have largely bifurcated into efficient retrieval and agentic reasoning. Retrieval-Augmented Generation (RAG) approaches~\citep{DBLP:journals/corr/abs-2502-01549, DBLP:journals/corr/abs-2508-01546, DBLP:journals/corr/abs-2505-23990} and frame selection strategies~\citep{DBLP:conf/iclr/YuJWCJZXSZWZS25, DBLP:conf/cvpr/YeWSCDEBN0FWL25, DBLP:journals/corr/abs-2502-06428} attempt to manage extreme token lengths by isolating key visual information. Simultaneously, agent-based frameworks~\citep{DBLP:conf/eccv/WangZZY24, DBLP:journals/corr/abs-2412-10471} and tool-augmented reinforcement learning methods~\citep{DBLP:journals/corr/abs-2508-04416, DBLP:journals/corr/abs-2506-10821} have introduced interactive reasoning to navigate video content. Despite these innovations, evaluations remain constrained, as they often overlook the high-level narrative progression essential for comprehending complex, long-form content such as TV series.

In this paper, we present~\textsc{SagaQA}, an abstractive question answering benchmark designed to evaluate multi-hop narrative reasoning over full-length TV series.~\textsc{SagaQA} addresses key limitations of prior datasets by introducing a \hlyfig{\emph{multi-video}}, \hlblue{\emph{multi-hop}} and \hlgrfig{\emph{multi-modal}} benchmark, where inputs are substantially longer and narratively more complex~(see Table~\ref{tab:dataset_comparison}). Each question targets a window of 20 consecutive episodes—approximately 20 hours of video—requiring models to perform ``big-hop'' reasoning to connect distant events from the same narrative chain. On average, answering a question from our dataset involves four reasoning hops, with relevant events potentially separated by up to 20 episodes. Therefore,~\textsc{SagaQA} constitutes a particularly challenging benchmark, especially in comparison to prior work that does not require the same depth of reasoning over complex narrative video content~(see Table~\ref{tab:dataset_comparison})—such as egocentric video understanding~\citep{DBLP:conf/cvpr/YangLGDZZWZXWOL25} or open-world internet video browsing~\citep{liang2026videobrowseragenticopenwebvideo}.

The dataset is constructed using the pipeline in Figure~\ref{fig:pipeline}. The resulting question-answer (QA) pairs undergo LLM-based filtering to ensure they satisfy the intended multi-hop reasoning criteria.

We study the performance of existing baselines on this benchmark including video planning methods, which we categorize into three classes: \emph{Parallel}, \emph{Sequential}, and \emph{Hybrid}. Our results show that hybrid planners consistently outperform the other approaches, achieving more complete and accurate episode grounding by combining broad exploration of candidate video segments with focused reasoning over the most relevant ones. By moving evaluations toward long-term, event-driven comprehension,~\textsc{SagaQA} provides valuable insights into the current limitations and future directions of multi-step reasoning in long-form video understanding.

Our contributions are as follows:
\begin{itemize}
\item We introduce~\textsc{SagaQA}, a novel benchmark specifically designed for evaluating long-form, multi-hop reasoning over TV series, addressing the gap between short-term visual perception and full-episode narrative understanding.
\item We systematically analyze video planning strategies, introducing the distinction between \emph{Parallel}, \emph{Sequential}, and \emph{Hybrid} Planners.
\item We demonstrate that hybrid planners achieve superior performance by balancing efficient exploration with deep multi-hop reasoning, providing insights for future research in long-form video understanding.
\end{itemize}

\begin{table*}
\center
\small
\setlength{\tabcolsep}{2pt} 
\resizebox{0.90\textwidth}{!}{
\begin{tabularx}{\textwidth}{X c c c c c c c l}
\toprule
\textbf{Dataset} & \textbf{Domain} & \textbf{Duration} & \textbf{M-Vid} & \textbf{M-Hop} & \textbf{Hop Length} & \textbf{Reasoning} & \textbf{M-Mod} & \textbf{\#Vid.} \\ 
\midrule

\multicolumn{1}{c}{\textbf{Long Video Understanding}} & & & & & & & \\

EgoSchema~\citep{DBLP:conf/nips/MangalamAM23} & Egocentric & 3min & \xmark & \xmark & -- & -- & \xmark & 5,063 \\

Video-MME~\citep{DBLP:journals/corr/abs-2405-21075} & Open Domain & 1--60min & \xmark & \xmark & -- & -- & \cmark & 900 \\

LVBench~\citep{DBLP:journals/corr/abs-2406-08035} & Open Domain & Up to 2h & \xmark & \xmark & -- & -- & \xmark & 103 \\

\noalign{\vskip 0.5ex}
\cdashline{1-9}[2pt/2pt]
\noalign{\vskip 0.5ex}

\multicolumn{1}{c}{\textbf{Multi-Hop Reasoning}} & & & & & & & \\

CausalStep~\citep{DBLP:journals/corr/abs-2507-16878} & Open Domain & 7min & \xmark & \cmark & \redd{sec--min} & \greenn{Narrative} & \xmark & 100 \\

Video-Holmes~\citep{DBLP:journals/corr/abs-2505-21374} & Short Films & 1--5min & \xmark & \cmark & \redd{sec--min} & \greenn{Narrative} & \cmark & 270 \\

VRBench~\citep{DBLP:journals/corr/abs-2506-10857} & Open Domain & 1.6h & \xmark & \cmark & \orangee{Up to 40min+} & \greenn{Narrative} & \cmark & 960 \\

\noalign{\vskip 0.5ex}
\cdashline{1-9}[2pt/2pt]
\noalign{\vskip 0.5ex}

\multicolumn{1}{c}{\textbf{Ultra-Long Video Reasoning}} & & & & & & & \\

Video-BrowseComp~\citep{liang2026videobrowseragenticopenwebvideo} & Open Web & hours & \cmark & \cmark & \redd{Search-Based} & \redd{Fact-Finding} & \cmark & Web \\

Ego-LifeQA~\citep{DBLP:conf/cvpr/YangLGDZZWZXWOL25} & Egocentric & 44.3h & \xmark & \cmark & \greenn{Up to 24h+} & \redd{Temporal} & \cmark & 6 \\

\midrule

\textit{SagaQA (ours)} & TV Series & 20h & \cmark & \cmark & \greenn{Up to 20h} & \greenn{Narrative} & \cmark & 525 \\

\bottomrule
\end{tabularx}
}
\caption{\textbf{Comparison of \textsc{SagaQA} with Existing Video Reasoning Benchmarks.} M-Vid: Multi-Video, M-Hop: Multi-Hop, M-Mod: Multi-Modal, \#Vid.: Number of Videos.}
\label{tab:dataset_comparison}
\end{table*}

\section{Related Work}

\paragraph{Long Video Understanding}
To evaluate the long-form understanding and reasoning capabilities of Video Large Language Msodels (Video-LLMs), several comprehensive benchmarks have been introduced~\citep{DBLP:conf/nips/MangalamAM23, DBLP:journals/corr/abs-2406-08035, DBLP:journals/corr/abs-2405-21075, DBLP:journals/corr/abs-2406-04264, DBLP:conf/cvpr/0002WH00LWX0L0024}. These are categorized as \emph{Long Video Understanding} datasets in Table \ref{tab:dataset_comparison}. To tackle these benchmarks, existing methods typically rely on query-guided retrieval to extract relevant information (often specific frames) from the entire video. This is generally achieved through zero-shot extraction~\citep{DBLP:journals/corr/abs-2504-17447, DBLP:journals/corr/abs-2406-09396, DBLP:journals/corr/abs-2405-19209}, tailored pretraining objectives~\citep{DBLP:conf/nips/Yu0YB23, DBLP:conf/iclr/YuJWCJZXSZWZS25, DBLP:conf/eccv/KorbarXTZT24}, or agentic frameworks~\citep{DBLP:conf/eccv/WangZZY24, DBLP:journals/corr/abs-2412-10471}.

\paragraph{Multi-Hop Reasoning and Agentic Approaches}
As video understanding has shifted from simple visual recognition to complex multi-hop reasoning, models are now required to logically connect disparate temporal events~\citep{DBLP:journals/corr/abs-2506-10857, DBLP:journals/corr/abs-2505-21374, DBLP:journals/corr/abs-2507-16878}. The corresponding datasets are categorized as \emph{Multi-Hop Reasoning} in Table \ref{tab:dataset_comparison}. To bridge the gap between raw visual recognition and high-level understanding, Chain-of-Thought (CoT) methodologies have become essential~\citep{DBLP:conf/cvpr/HanHSZSZZ0L025}, alongside a growing line of agentic approaches~\citep{DBLP:journals/corr/abs-2504-04471, DBLP:journals/corr/abs-2508-04416, DBLP:journals/corr/abs-2506-10821, DBLP:journals/corr/abs-2509-23652}. However, recent research highlights significant limitations in agentic video understanding. Autonomous models frequently suffer from \emph{reasoning drift}, where their logic diverges from the original query over time, a problem often compounded by \emph{noise contamination} from imperfect visual tools~\citep{DBLP:journals/corr/abs-2504-04471, DBLP:journals/corr/abs-2601-06943}. Motivated by the above limitations, our study compares and evaluates different agentic strategies under highly complex reasoning constraints from our~\textsc{SagaQA} dataset.

\paragraph{Reasoning on Ultra-Long Video Narratives}
Recent work on ultra-long video understanding has heavily focused on egocentric settings, where untrimmed recordings span from hours to entire days of human activity~\citep{DBLP:conf/nips/ChandrasegaranG24, DBLP:conf/cvpr/YangLGDZZWZXWOL25}. Because end-to-end models struggle at this scale, recent methods introduce modular reasoning paradigms, such as Chain-of-Tool-Thought~\citep{DBLP:journals/corr/abs-2506-13654, DBLP:journals/corr/abs-2601-18157}, to decouple high-level planning from tool execution. A parallel line of work explores similar agentic approaches for large-scale video browsing over the open web~\citep{liang2026videobrowseragenticopenwebvideo}.

While these prior works push the temporal boundaries of video reasoning, Table~\ref{tab:dataset_comparison} shows they primarily emphasize \emph{Fact-Finding} over the web or \emph{Temporal} understanding such as tracking when a person last interacted with an object in egocentric videos, as seen in EgoLifeQA~\citep{DBLP:conf/cvpr/YangLGDZZWZXWOL25}. In contrast, the \textsc{SagaQA} dataset differs fundamentally from those previous works by requiring models to uncover deep \emph{Narrative} structures and causal dependencies that span multiple TV episodes (4.2 episodes on average). Because these \emph{Narrative} dependencies are widely separated in time, they span substantially longer temporal windows than those reported in prior works (see Table~\ref{tab:dataset_comparison}), presenting a novel challenge for video agents.

\section{Benchmark Construction}

\subsection{Dataset Description}
\label{subsec:distrib_data}

Our dataset,~\textsc{SagaQA}, contains 200 QA pairs spanning~4 years of the TV show~\textit{As the World Turns}~(2007-2010), providing a sufficient resource for controlled evaluation of multi-hop reasoning across extended video narratives. Each sample consists of a question~$Q$ associated with a set of 20 consecutive episodes~$\mathcal{E}_{20}$, for which the corresponding videos are provided. The task is framed as abstractive question answering; the ground-truth annotations include the answer~$A$ to the question and the subset~$\mathcal{E}^*$ of episodes that are relevant for answering it. Formally, a sample can be represented as~($Q$, $\mathcal{E}_{20}$, $A$, $\mathcal{E}^*$).

We impose three key constraints on the QA pairs in the~\textsc{SagaQA} dataset:

\begin{itemize}
    \item \hlyfig{\textbf{Multi-Episode}} The questions cannot be answered using a single episode.
    \item \hlblue{\textbf{Multi-Hop}} The questions must integrate information from multiple audio-visual segments through a series of dependent reasoning steps that progressively build toward the final answer.
    \item \hlgrfig{\textbf{Multi-Modal}} The questions cannot be answered from dialogue alone and require analyzing visual elements from the video.
\end{itemize}

In Appendix~\ref{app:samples_sagaQA}, we provide three dataset samples with manual annotations of the multi-hop reasoning path, the relevant episodes, and the multimodal elements that contribute to each answer.

\begin{figure}[htbp]
    \centering
    \includegraphics[width=0.45\textwidth]{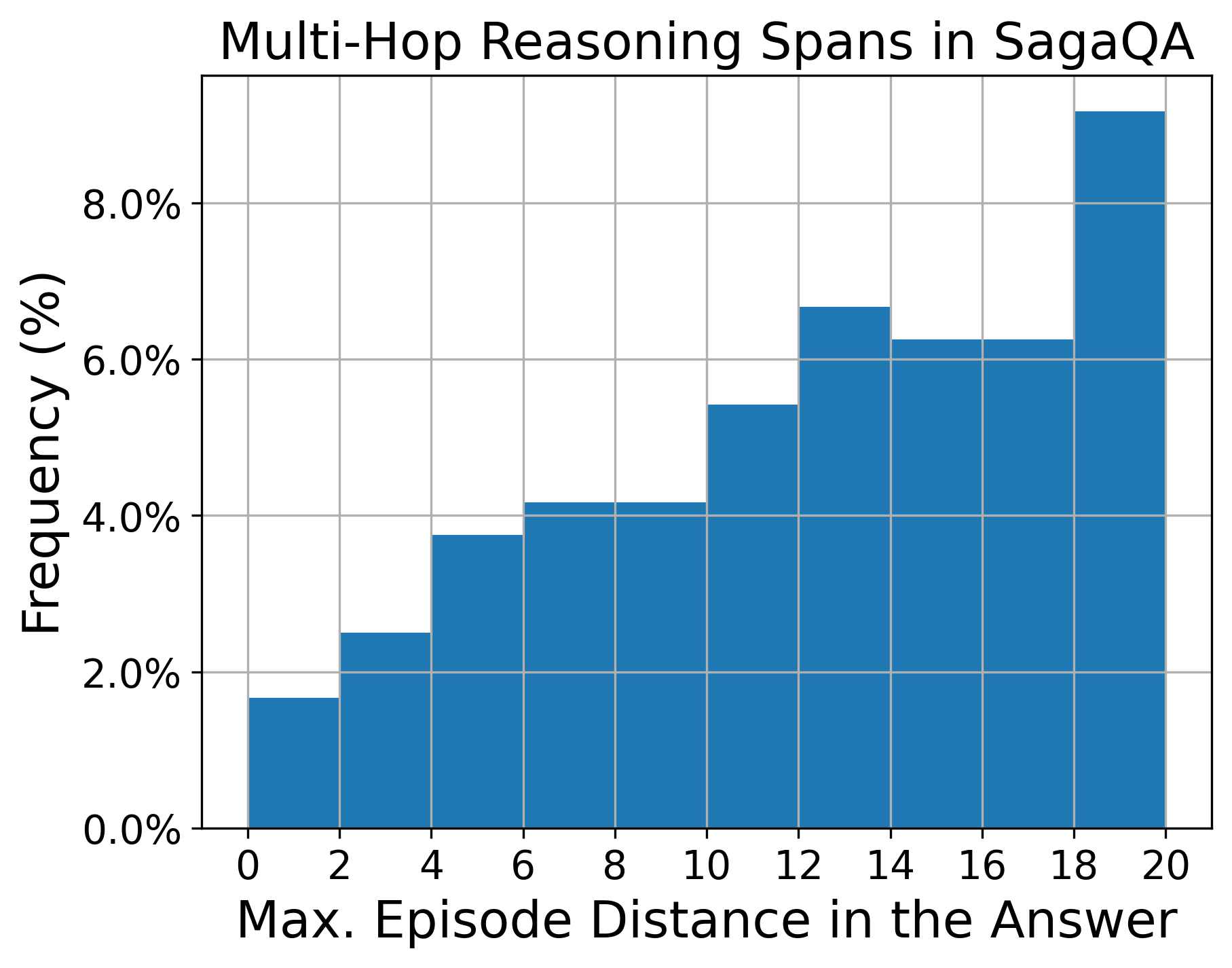}
    \includegraphics[width=0.45\textwidth]{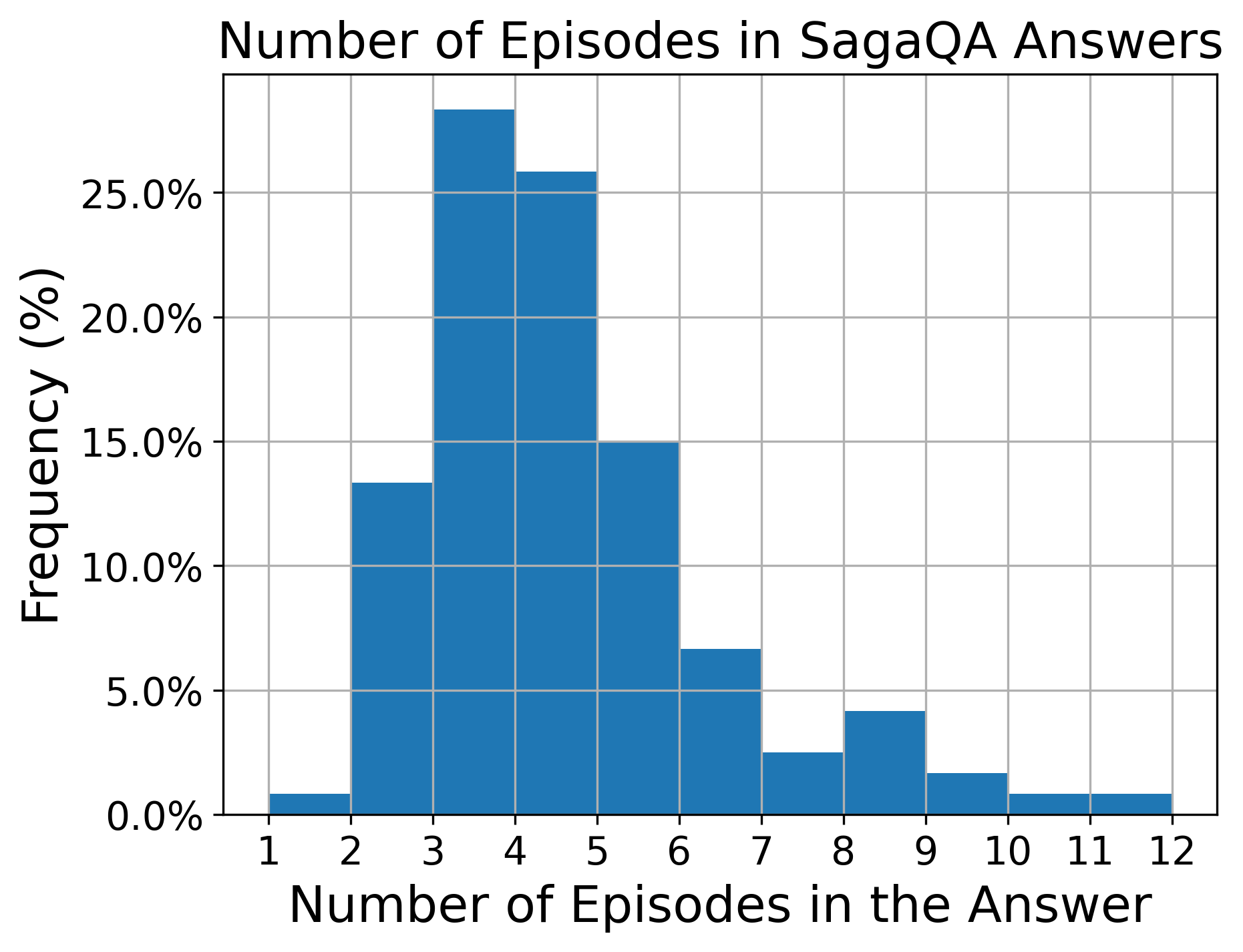}
    \caption{\textbf{Statistics of~\textsc{SagaQA}.} We provide the distribution of the maximum distance between two episodes involved in the multi-hop reasoning sequence as well as the distribution of the number of episodes involved in answering per question for all the questions in our dataset.}
    \label{fig:stats_SagaQA}
    \vspace{-0.3cm}
\end{figure}

\subsection{Dataset Statistics} \label{sec:distribution}
On average, each question involves approximately~$4.2$ episodes and rarely exceeds~7 episodes, reflecting the high number of hops required for the questions~(Figure~\ref{fig:stats_SagaQA}).

The size of each hop, however, varies substantially within the 20-episode window. As illustrated by the distribution in Figure~\ref{fig:stats_SagaQA}, most questions in our dataset exhibit relatively long dependency chains, with a median temporal length of~13 episodes and a standard deviation of~$5.4$ episodes. Given that each episode has a fixed duration of one hour, the resulting reasoning chains span unusually long temporal intervals compared to other datasets in the literature~(see Table~\ref{tab:dataset_comparison}).

\subsection{QA Pairs Generation} \label{sec:qa_pair_generation}
\label{subsec:generation_pipeline}
We rely on existing annotations in the SummScreen3D dataset~\citep{DBLP:conf/eacl/PapalampidiL23}. SummScreen3D is a video-to-text summarization dataset of 5421 episodes each from popular soap operas like~\textit{As the World Turns}. It includes full transcripts, videos and multiple summaries for each episode. The summaries in SummScreen3D are also highly multimodal as they contain information referring to both the episode video and transcripts. This is crucial for generating our multimodal QA pairs.

For building our~\textsc{SagaQA} dataset we focus on the TV show~\textit{As the World Turns} as we notice that the TV show requires a proper multimodal understanding beyond simply the dialogue. Each QA pair in our dataset spans over~20 consecutive episodes, corresponding to about~20 hours of videos.

We detail below, and illustrate in~Figure~\ref{fig:pipeline}, the iterative process for generating each sample~($Q$,~$\mathcal{E}_{20}$,~$A$,~$\mathcal{E}^*$) in~\textsc{SagaQA}. All prompts are given in Appendix~\ref{app:qa_pair_generation}.

\begin{itemize}
    \item \textbf{Step~1:} Identify all the characters listed within the 20-episode context window~$\mathcal{E}_{20}$. We denote as~$\mathcal{C}$, the list of all extracted characters from the corresponding~20 episode summaries in SummScreen3D.
    \item \textbf{Step~2:} Given a character in~$\mathcal{C}$ and the~20 episode summaries, we generate a QA pair for that character. The prompt given to Gemini 2.5 Flash~\citep{DBLP:journals/corr/abs-2507-06261} precisely states the constraints on the generated pair:~\hlyfig{\emph{multi-episode}}, \hlblue{\emph{multi-hop}} and \hlgrfig{\emph{multi-modal}}~(see Appendix~\ref{app:qa_pair_generation}). We denote as~$\mathcal{C}_{\text{QA}}$ the list of characters involved in the QA pair returned by the LLM.
    \item \textbf{Step~3:} We remove~$\mathcal{C}_{\text{QA}}$ from the set of all the characters~$\mathcal{C}$.
    \item \textbf{Step~4:} We repeat from \textbf{Step~2} until~$\mathcal{C}$ is empty.
    \item \textbf{Step~5:} We finally post-process all generated QA pairs to retain only the ones that are truly multi-hop. We use the prompt in Appendix~\ref{app:multi_hop_filtering}.
\end{itemize}

\begin{figure*}[hbt]
\begin{center}
\includegraphics[width=0.80\textwidth]{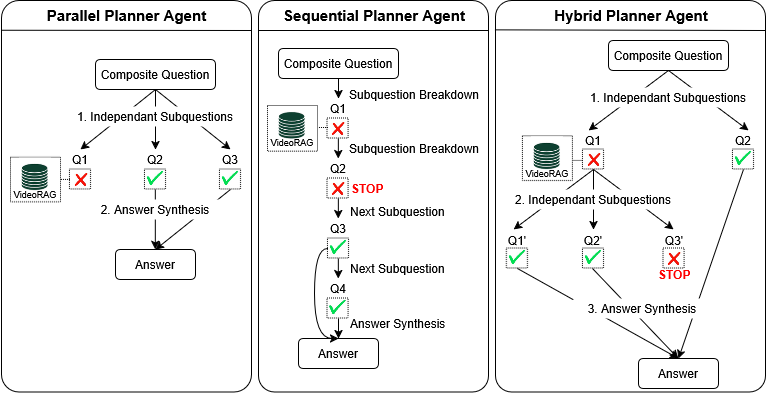}
\end{center}
\caption{\textbf{Existing strategies for video planning.} From left to right:~1) the \emph{Parallel Planner} splits a composite question into a set of independent subquestions;~2) the \emph{Sequential Planner} generates subquestions one at a time, progressively working toward an answer;~3) the \emph{Hybrid Planner} combines the two previous strategies by iteratively producing independent requests to the tools while eventually constructing an answer.~\textcolor{red}{\textbf{STOP}} denotes the end of a reasoning trajectory executed by an agent. The green checkmark indicates that the tools successfully produced an answer, while the red cross indicates a failure.}
\label{fig:agentic_framework}
\vspace{-0.3cm}
\end{figure*}

\subsection{Human Verification}
\label{subsec:validation_pipeline}

We validate our data generation process using two human evaluators: one co-author and one external researcher. Each evaluator is provided with the same set of~25 QA pairs.

To support the evaluation process, annotators are allowed to consult the output of an external LLM (Gemini 2.5 Flash), but only through a restricted set of prompts to ensure consistent and reliable responses. Specifically, for each episode~$E_i$, evaluators may access the LLM-generated answer~$A_i$ to the question, where~$A_i$ is directly produced from the respective annotated summary of the episode for more robustness.

We provide the following guidelines to assess each QA pair with respect to our criteria:

\begin{itemize}
    \item \textbf{Multi-Episode} Verify that no single episode-level answer~$A_i$ contains sufficient information to derive the final answer~$A$.
    \item \textbf{Episode List} Identify the subset of episodes~$\mathcal{E}_{\text{human}}$ required to construct the final answer~$A$. This corresponds to selecting all episodes~$E_i$ whose associated answers~$A_i$ contribute to~$A$. We then compute Precision, Recall, and F1 scores between the episode list~$\mathcal{E}^*$ in~\textsc{SagaQA} and the annotated reference~$\mathcal{E}_{\text{human}}$.
    \item \textbf{Multi-Hop} Assess whether the different elements required to answer question~$Q$ are logically connected and interdependent.
\end{itemize}

We report below the average results found by both annotators.

\begin{itemize}
    \item \textbf{Multi-Episode} 95\% of the answers in~\textsc{SagaQA} involves combining information from different episodes.
    \item \textbf{Episode List} The average Precision, Recall and F1 scores for the episode list annotation of each sample are~87\%,~93\% and~90\% respectively. We obtain a Jaccard similarity of~86\% between the episode lists identified by the two annotators.
    \item \textbf{Multi-Hop} The step for filtering multi-hop questions (\textbf{Step~5} of our data generation,         Section~\ref{sec:qa_pair_generation}) achieved a precision of~82\%, recall of~91\%, and an F1 score of~86\%. Cohen's~$\kappa$ was~0.61, indicating substantial agreement between the two annotators.
\end{itemize}

\section{Multi-Hop Video Reasoning with Agentic Planners}
\label{sec:comparison_models}

\subsection{Video Planning}

Long TV shows are too large for end-to-end processing by VLMs. Additionally, multi-hop questions often require step-by-step reasoning, making static single-turn retrieval highly ineffective~\citep{DBLP:conf/emnlp/PressZMSSL23}.
Multi-hop reasoning is commonly addressed through planning strategies in the literature. A planner aims to answer a composite question, i.e., a question that requires both independent (parallel) and dependent (multi-hop) reasoning steps to produce a final answer. Typically, it generates a set of queries that are executed by external tools and whose outputs are aggregated to construct a global answer. Video planners can be categorized into three types (Figure~\ref{fig:agentic_framework}).

\textbf{Parallel Planner} A planner that decomposes a composite question into a set of simpler, independent subquestions, which are then answered in parallel~\citep{DBLP:journals/corr/abs-2601-18157}.

\textbf{Sequential Planner} A planner that answers a composite question through a sequence of steps~\citep{DBLP:journals/corr/abs-2508-04416, DBLP:journals/corr/abs-2506-10821, DBLP:conf/eccv/WangZZY24, DBLP:journals/corr/abs-2506-13654}. At each step (Figure~\ref{fig:agentic_framework}), the agent selects an action from a predefined set, generates a subquestion, and invokes the corresponding tools. Overall, the sequential planner issues queries one at a time, progressively building toward the final answer.

\textbf{Hybrid Planner} We introduce the hybrid planner as a way to combine the strengths of both parallel and sequential approaches. Like the sequential planner, the agent selects an action at each step. However, similar to the parallel planner, it generates and executes multiple independent queries simultaneously at each step.
This design significantly reduces the length of reasoning chains. It mitigates the main limitations of sequential planning:~1) reasoning traces tend to drift and accumulate noise as depth increases, and~2) low cost efficiency due to the inability to execute queries in parallel. At the same time, it preserves the flexibility needed to handle multi-hop dependencies, which purely parallel planners fail to capture.

\subsection{Video Planners Implementation}

In this section, we describe each studied video planning strategy and provide full implementation details for all methods in Appendix~\ref{app:baseline_prompting}. In our experiments, we use the VideoRAG framework~\citep{DBLP:journals/corr/abs-2502-01549} as the only tool for retrieving video information. In the VideoRAG implementation, an LLM judges the confidence of the retrieved video segments. If all retrieved segments are assigned low confidence, the system reports back to the planner that no answer was found for the query~(\xmark).

\textbf{Parallel Planner + VideoRAG} We study two variants of this approach. The first implementation~(Prompt~1) extracts keywords from the question, groups them by similarity, and generates one subquestion per group to query the VideoRAG. In the second implementation~(Prompt~2), an LLM performs an end-to-end decomposition of the original composite question into subquestions.

\textbf{Sequential Planner + VideoRAG} In this setting, we choose the set of actions from among (\texttt{BREAKDOWN}, \texttt{MOVE\_ON}).

The \texttt{BREAKDOWN} action is selected when the previous VideoRAG query fails to yield a satisfactory answer~(\xmark), prompting the agent to further decompose the query.

The \texttt{MOVE\_ON} action is selected either when the answer to the previous subquestion was satisfactory~(\cmark) or when repeated attempts fail to produce useful results~(\textcolor{red}{\textbf{STOP}}). In both cases, the agent generates the next subquestion, targeting a different aspect of the original query.

\textbf{Hybrid Planner + VideoRAG} Similar to the sequential planner, the agent dynamically selects between the \texttt{BREAKDOWN} and \texttt{MOVE\_ON} actions. However, in this setting, the \texttt{BREAKDOWN} action is executed in parallel: instead of generating subquestions sequentially, the agent produces a set of independent subquestions that are queried concurrently to the VideoRAG.

\begin{table*}
\center
\setlength{\tabcolsep}{2pt}
\resizebox{0.90\textwidth}{!}{
 \begin{tabular}{@{}lcccc|ccc@{}}
\toprule
& r1 & r2 & rlsum & METEOR & Precision & Recall & F1 \\
\midrule
\multicolumn{1}{c}{\textbf{Non-Planning}} & & & & & & & \\
\multicolumn{1}{l}{TextRAG} & 36.30 & 6.51 & 33.09 & 23.49 & 28.13 & 11.14 & 14.92 \\
\multicolumn{1}{l}{VideoRAG} & 36.71 & 6.72 & 33.27 &~\textbf{24.42} & 36.91 & 30.46 & 28.54 \\
\noalign{\vskip 0.5ex}
\cdashline{1-8}[2pt/2pt]
\noalign{\vskip 0.5ex}
\multicolumn{1}{c}{\textbf{Parallel Planners}} & & & & & & & \\
\multicolumn{1}{l}{EGAgent~(Qwen3-30B-Instruct)} & 33.01 & 4.25 & 29.41 & 20.38 & 38.89 & 33.33 & 32.86 \\
\multicolumn{1}{l}{\textit{Parallel Planner (Prompt 1) + VideoRAG}} & 35.00 & 5.93 & 31.93 & 23.42 &~\textbf{50.03} & 26.35 & 30.88 \\
\multicolumn{1}{l}{\textit{Parallel Planner (Prompt 2) + VideoRAG}} & 36.06 & 6.15 & 32.98 & 24.10 & 48.06 & 30.93 & 34.26 \\
\noalign{\vskip 0.5ex}
\cdashline{1-8}[2pt/2pt]
\noalign{\vskip 0.5ex}
\multicolumn{1}{c}{\textbf{Sequential Planners}} & & & & & & & \\
\multicolumn{1}{l}{VideoExplorer (7B)} & 27.18 &~\textbf{10.67} & 23.89 & 18.94 & 33.33 & 10.42 & 15.83 \\
\multicolumn{1}{l}{\textit{Sequential Planner + VideoRAG}} &~\textbf{36.94} & 6.67 &~\textbf{33.35} & 24.22 & 46.76 & 30.43 & 32.68 \\
\noalign{\vskip 0.5ex}
\cdashline{1-8}[2pt/2pt]
\noalign{\vskip 0.5ex}
\multicolumn{1}{c}{\textbf{Hybrid Planners}} & & & & & & & \\
\multicolumn{1}{l}{\textit{Hybrid Planner + VideoRAG}} & 35.99 & 6.32 & 32.58 & 24.07 & 45.47 &~\textbf{44.11} &~\textbf{41.02} \\

\bottomrule
\end{tabular}
}
\caption{\textbf{Results on the test set with Qwen3-30B-Instruct.} We report ROUGE-1 (r1), ROUGE-2 (r2), ROUGE-Lsum (rlsum) and METEOR. We also include episode grounding metrics with the Precision, Recall and F1 scores. Best results are in \textbf{bold}.}
\label{tab:main-results-MFactSum}
\end{table*}

\subsection{Implementation Details} \label{subsec:implementation_details}

Unlike the original implementation of VideoRAG, which relies on Gemma-2-9B~\citep{DBLP:journals/corr/abs-2408-00118}, we use Qwen3-30B-Instruct~\citep{DBLP:journals/corr/abs-2505-09388} or Mistral-Small-3.1-24B~\citep{mistralsmall31_2025} as the language model for both planning and final answer generation. In the prompt to all video planners, we always fix the length of the final answer to~200 words~(see Appendix~\ref{app:answer_synthesis_prompt}). We limit the maximum depth of the reasoning sequence to~15 in the sequential planner as we notice that reasoning beyond that limit almost always ends up looping over the same reasoning steps. For the hybrid planner, we limit the depth of the reasoning tree to~2.

\section{Experiments}
\label{sec:experiments}

\subsection{Evaluation Metrics} \label{subsec:clip_selection_metrics}

\paragraph{Text Generation Metrics} We report ROUGE-1 (r1), ROUGE-2 (r2), and ROUGE-Lsum (rlsum) using the python-rouge package, as well as METEOR scores computed with the \texttt{meteor\_score} function from \texttt{nltk.translate}.

\paragraph{Episode Grounding Metrics} We report Precision, Recall and F1 to assess if the episodes retrieved by the system match the annotations in~\textsc{SagaQA}.

\subsection{Comparative Baselines}

We group the baselines into planning and non-planning methods in Table~\ref{tab:main-results-MFactSum}. Among the planning methods, VideoExplorer is a trained sequential planner, while EGAgent follows a parallel planning strategy. We exclude agentic baselines that focus solely on visual information~\citep{DBLP:conf/eccv/WangZZY24, DBLP:journals/corr/abs-2511-20785}, as our dataset requires multimodal understanding of audio and visual signals. However, we include \textsc{TextRAG} as a single-modality baseline to specifically assess the role of multimodal reasoning in \textsc{SagaQA}. Finally, given
the extreme length of TV show content, direct end-to-end reasoning with Video-LLMs is computationally infeasible.

\textbf{VideoRAG}~\citep{DBLP:journals/corr/abs-2502-01549} Retrieval Augmented Generation (RAG) implementation for scalable understanding and generation over hours of videos along with audio.

\textbf{TextRAG} We modify the above VideoRAG implementation to perform query matching solely based on the automatically transcribed subtitles.

\textbf{VideoExplorer~(7B)}~\citep{DBLP:journals/corr/abs-2506-10821} VideoExplorer is an agentic long-video understanding framework that is trained to iteratively plan subquestions, performs temporal grounding, and conducts task-driven video exploration to enable scalable reasoning over extended video content. We let the agent process both the video and transcribed subtitles with Whisper-Small~\citep{DBLP:conf/icml/RadfordKXBMS23} provided in the~\texttt{whisper} Python package.

\textbf{EGAgent~(with Qwen3-30B-Instruct)}~\citep{DBLP:journals/corr/abs-2601-18157} EGAgent is a framework for long video understanding that leverages entity-centric scene graphs and hybrid multimodal retrieval for multi-hop reasoning.

\subsection{Results} \label{subsec:clip_results}

We split the dataset into two subsets, using~50 QA pairs for validation and~150 for testing. For each tested baseline, we identify the set of episodes used to ground its generated answers. This predicted episode set is then compared to the annotations in~\textsc{SagaQA} to compute the Precision, Recall, and F1 scores reported in Table~\ref{tab:main-results-MFactSum}. For methods relying on VideoRAG, this set corresponds to the episodes from which high-confidence video segments are retrieved; for VideoExplorer and EGAgent, we instead parse the model outputs to extract the episodes used to support their final answers.

\paragraph{The Importance of Multimodality} We notice that all the episode grounding metrics drop when using TextRAG instead of VideoRAG, suggesting that visual information plays a meaningful role in answering the questions in~\textsc{SagaQA}.

\paragraph{Hybrid Planning for Better Episode Grounding} The hybrid planner achieves substantially stronger episode grounding than the other evaluated video planners as indicated by higher F1 scores in Table~\ref{tab:main-results-MFactSum}. This improvement indicates that it effectively combines the strengths of parallel and sequential strategies, leveraging both broad exploration and deep reasoning (see Section~\ref{subsec:discussion}).

Importantly, these observations hold consistently when replacing Qwen3-30B-Instruct with Mistral-Small-3.1-24B in all VideoRAG-based methods (see Appendix~\ref{app:experiments}).

\section{Discussion} \label{subsec:discussion}

\paragraph{Exploration–Exploitation Trade-Off} Our experiments reveal a systematic trade-off across planners. Parallel planners issue multiple queries in a single step, enabling broad coverage of the search space but limiting their ability to perform iterative, multi-hop refinement. In contrast, sequential planners refine queries over multiple steps, but tend to over-exploit early high-confidence directions, repeatedly issuing similar queries and failing to adequately cover alternative directions. Our hybrid planner mitigates both limitations and achieves significantly higher F1 scores~(Table~\ref{tab:main-results-MFactSum}). To illustrate this behavior, we compare, in Appendix~\ref{app:reasoning_trace}, the reasoning traces of our hybrid planner with those of two sequential planners, highlighting the latter’s tendency to repeatedly issue similar queries instead of effectively expanding the search space.

\paragraph{Limitations in Answer Synthesis}
Even when the planner retrieves the relevant video segments required to answer the question, effectively exploiting this evidence to produce a high-quality answer remains challenging. While episode grounding metrics vary significantly across the studied settings, abstractive metrics—such as ROUGE and METEOR—reported in Table~\ref{tab:main-results-MFactSum} remain largely unchanged. This discrepancy suggests that current methods are limited not in retrieval, but in downstream information synthesis. In particular, even when the correct episode is identified, the planner may fail to capture the precise narrative pivot (e.g., a subtle facial expression or a key visual cue), since the agentic planner processes video segments as isolated units rather than as parts of a coherent temporal arc. This highlights the need for more structured post-retrieval reasoning and aggregation mechanisms.

\paragraph{Toward Structured and Hybrid Planning Strategies}
The strong performance of the hybrid planner underscores the benefits of combining complementary reasoning paradigms. The parallel component acts as a broad exploration mechanism, identifying a diverse set of candidate video segments, while the sequential component serves as a knowledge refiner, ensuring that multi-hop dependencies are coherently connected. This division of roles suggests that future long-form agents should move beyond single-trajectory reasoning and instead adopt more structured exploration strategies, such as Tree-of-Thought~\citep{DBLP:conf/nips/YaoYZS00N23}, tree-based or graph-based planning frameworks~\citep{DBLP:journals/corr/abs-2512-03500}. Further improvements could be achieved through targeted training, as well as the integration of specialized retrieval and grounding tools to enhance both coverage and reasoning consistency.

\section{Conclusion}

In this paper, we introduced~\textsc{SagaQA}, a novel benchmark designed to push the boundaries of long-form video understanding from local perception to high-level narrative comprehension. By focusing on multi-hop reasoning across 20-episode windows—representing approximately 20 hours of continuous video—\textsc{SagaQA} challenges models to connect disparate events, character arcs, and multimodal cues that span significant temporal distances.

Our systematic evaluation of various video planning strategies reveals a critical~\emph{exploration-exploitation} trade-off in long-form reasoning. While parallel planners provide broad coverage of the search space, they often lack the depth required for complex causal chains. Conversely, sequential planners excel at refinement but are prone to reasoning drift and repetitive loops. We demonstrate that hybrid planners, which combine parallel exploration and sequential refinement, achieve superior performance, particularly in grounding answers to the correct episodes.

Our results suggest that, beyond retrieval, a key bottleneck lies in effectively synthesizing evidence across extended narratives. Even when relevant episodes are correctly identified, current models struggle to consistently aggregate this information into accurate answers. By releasing~\textsc{SagaQA} and our evaluation framework, we aim to facilitate further research in long video reasoning.

\section*{Limitations}

On the evaluation side, the proposed task-specific metrics—Episode Precision, Recall, and F1—are both challenging and sufficiently discriminative. However, the current annotations in~\textsc{SagaQA} could be further manually refined by experts by grounding the answers at the clip level rather than entire episodes. These finer-grained annotations would allow for a more precise and informative evaluation than episode-level assessment.

The benchmark is built exclusively from TV shows~(Section~\ref{sec:qa_pair_generation}). Soap operas have very specific narrative structures (e.g., strong cross-episode narrative continuity, dialogue-driven storytelling, and recurring characters). Models might overfit to this genre, and the results may not generalize to other video domains such as documentaries or instructional videos.

While~20 hours of video is long-form, the benchmark uses a fixed window of~20 consecutive episodes~(Section~\ref{subsec:distrib_data}). In actual TV series, character arcs and plot hops can span hundreds of episodes over several years.~\textsc{SagaQA} does not yet test~\emph{ultra-long} memory that spans the entire lifecycle of a TV series.

\section*{Acknowledgments}
We thank the anonymous reviewers for their feedback. We are also grateful to Parth Sakhare for his assistance with the human evaluation of~\textsc{SagaQA}. This research is supported by the National Research Foundation, Prime Minister’s Office, Singapore under its Campus for Research Excellence and Technological Enterprise (CREATE) programme. Any opinions, findings and conclusions or recommendations expressed in this material are those of the authors and do not reflect the views of the National Research Foundation, Singapore.

\bibliography{custom}
\bibliographystyle{acl_natbib}

\appendix

\section{\textsc{SagaQA} Examples} \label{app:samples_sagaQA}

We provide three different examples from~\textsc{SagaQA}. We also manually studied whether these samples satisfy the criteria defined for the dataset, namely~\hlyfig{\emph{multi-episode}}, \hlblue{\emph{multi-hop}} and \hlgrfig{\emph{multi-modal}}.

\begin{itemize}
    \item \hlyfig{\textbf{Multi-Episode}} The answers explicitly refer to multiple episodes. We highlight in~\hlyfig{YELLOW} references to the episodes.
    \item \hlblue{\textbf{Multi-Hop}} We manually annotate in~\hlblue{BLUE} the reasoning hops required to answer each of the questions below. The questions require~5,~4 and~3 hops respectively.
    \item \hlgrfig{\textbf{Multi-Modal}} After watching the episodes relevant to each of the questions, we manually highlight in~\hlgrfig{GREEN} the parts of the answer that directly refers to the video content.
\end{itemize}

\subsection{Example~1}

\begin{tcolorbox}[breakable, colback=white, colframe=gray!20, left=2pt,  coltitle=black,  title=\textbf{}]

\textbf{Question~$Q$}:\\
\hlblue{(Hop 1)} How did Midnight Sun vodka samples~\hlblue{(Hop 2)} at Carly's home~\hlblue{(Hop 3 + Hop 4)} lead to danger for her children,~\hlblue{(Hop 5)} and what was her decision about the business, given her past with an alcoholic father and the observable proof of the incident?\\

\textbf{Episodes to Answer~$\mathcal{E}^*$}: 9\textsuperscript{th}, 20\textsuperscript{th}, February and 4\textsuperscript{th} March 2009\\

\textbf{Answer~$A$}:\\
\hlblue{(Hop 1)} Carly, initially excited about her Midnight Sun vodka venture with Craig,
\hlblue{(Hop 2)} found herself in a dangerous situation when~\hlgrfig{Craig brought unlabeled product samples to her home.} \hlblue{(Hop 3)}~\hlgrfig{Her children, Parker and Liberty, mistook these for soda, drank them, and became drunk.}~\hlblue{(Hop 4)} This led to them ditching school and leaving a pizza in the oven, causing a fire and leaving her kitchen with~\hlgrfig{blackened appliances and counters.}\\
\hlblue{(Hop 5)} Upon seeing the damage and realizing she could have lost her son, Carly was deeply shocked and emotional. She revealed her childhood trauma of being terrorized by an alcoholic father, expressing her fear that Parker might follow a similar path. Despite this profound realization and the incident's severity, Carly initially decided not to shelve the Midnight Sun project, affirming her partnership with Craig. However, the event profoundly underscored her deep-seated fears about alcohol's impact on her family.

\end{tcolorbox}

\subsection{Example~2}

\begin{tcolorbox}[breakable, colback=white, colframe=gray!20, left=2pt,  coltitle=black,  title=\textbf{}]

\textbf{Question~$Q$}:~\hlblue{(Hop 1)} After Emily hired her to separate Casey and Alison,~\hlblue{(Hop 2 + Hop 3)} how did Jade romantically pursue Casey,~\hlblue{(Hop 4)} and how did he outwardly respond to her advances?\\

\textbf{Episodes to Answer~$\mathcal{E}^*$}: 6\textsuperscript{th} and 19\textsuperscript{th} February 2009\\

\textbf{Answer~$A$}:\\
\hlblue{(Hop 1)}~\hlyfig{On February 6th}, after being secretly hired by Emily to keep Alison and Casey apart,
\hlblue{(Hop 2)} Jade developed her own romantic interest in Casey.\\
\hlblue{(Hop 3)}~\hlyfig{On February 19th}, after Casey cancelled their date, Jade, wearing her new dress, went to Memorial Hospital. She found Casey in an~\hlgrfig{exam room}, where she~\hlgrfig{pulled the blinds, locked the door, and gave Casey an inviting kiss}.~\hlblue{(Hop 4)} However, Casey~\hlgrfig{pushed her away and made a break for the hall,} clearly rejecting her advance.

\end{tcolorbox}

\subsection{Example~3}

\begin{tcolorbox}[breakable, colback=white, colframe=gray!20, left=2pt,  coltitle=black,  title=\textbf{}]

\textbf{Question~$Q$}:\\
\hlblue{(Hop 1 + Hop 2)} How did Craig create a secret bond with Parker,~\hlblue{(Hop 3)} and how did it lead to Craig secretly funding Parker’s Metallica tickets?\\

\textbf{Episodes to Answer~$\mathcal{E}^*$}: 2\textsuperscript{nd}, 9\textsuperscript{th} and 30\textsuperscript{th} January 2009\\

\textbf{Answer~$A$}:\\
\hlblue{(Hop 1)} On \hlyfig{January 2nd},~\hlgrfig{Craig gave Parker his car keys} for Parker's 16th birthday, allowing him to drive to hockey practice. Parker, however, took Liberty on a joyride, and~\hlgrfig{they ended up stranded and making out}. Jack was furious at Craig for loaning the car.\\
\hlblue{(Hop 2)} A week later,~\hlyfig{on January 9th}, Parker and Liberty~\hlgrfig{were caught by Craig making out behind the bar at Metro}, where Parker was doing chores for Carly. Craig, still possessing a key, found them but covered for their presence,~\hlgrfig{rushing Liberty out the back door before Carly arrived}. This created a secret bond and leverage for Craig.\\
\hlblue{(Hop 3)} Later,~\hlyfig{on January 30th}, Parker needed \$200 for Metallica concert tickets, but Jack, facing financial strain, couldn't provide the money and reiterated his disapproval of Craig's influence. Craig, aware of Parker's desire for the tickets and having previously covered for him, gave Parker money to buy toys for Johnny, explicitly telling him to keep extra for his trouble.~\hlgrfig{Parker's eyes lit up}, and he used this money to buy the Metallica tickets, effectively having Craig fund his entertainment despite Jack's strong objections.

\end{tcolorbox}

\section{Prompts}

\subsection{QA Pair Generation}
\label{app:qa_pair_generation}

\subsubsection{Character Identification Prompt}

We provide the prompt below to identify all the characters mentioned in a single episode summary within the~20-episode context window~$\mathcal{E}_{20}$. The prompt is given to Qwen3-30B-Instruct.

\begin{itemize}
    \item {\color{red} <SUMMARY\_i>} is the groundtruth summary of episode~$E_i$ from the episode list~$\mathcal{E}_{20}$.
\end{itemize}

\begin{tcolorbox}[breakable, colback=white, colframe=gray!20, left=2pt,  coltitle=black,  title=\textbf{}]
{\color{red} <SUMMARY\_i>}\\

List the main characters actively participating in the above episode separated by a comma. I want only the characters who have a known first name.\\

Characters:
\end{tcolorbox}

\subsubsection{QA Pair Generation Prompt} \label{qa_pair_generation_prompt}

We provide the prompt below to generate the QA pairs in~\textsc{SagaQA}. The prompt is given to Gemini-2.5-Flash.

\begin{itemize}
\item {\color{blue} DATE\_EPISODE\_i} is the date of the episode~$E_i$.
\item {\color{red} <SUMMARY\_i>} is the groundtruth summary of episode~$E_i$.
\item {\color{blue} CHARACTER} is the character from the TV show around which the question must be built.
\end{itemize}

\begin{tcolorbox}[breakable, colback=white, colframe=gray!20, left=2pt,  coltitle=black,  title=\textbf{}]

Summary {\color{blue} DATE\_EPISODE\_1}:\\
{\color{red} <SUMMARY\_1>}\\

Summary {\color{blue} DATE\_EPISODE\_2}:\\
{\color{red} <SUMMARY\_2>}\\

\ldots\\

Summary {\color{blue} DATE\_EPISODE\_20}:\\
{\color{red} <SUMMARY\_20>}\\

Answer in the following way:
\begin{enumerate}
    \item Multimodal Multi-Episode Multi-Hop Question around character {\color{red} <CHARACTER>} involving multiple visual elements or actions (Do not provide the answer within the question, Keep your question clear and simple):
    \item Compact Question (Write the same question in about 30 words):
    \item Justification why the question is Multi-Hop. For each hop, explain how it relates to the previous hops and which reasoning type is involved (Bridge Entities (Compositional), Comparison \& Intersection, Causal \& Counterfactual, Temporal Reasoning, Multi-Step Quantitative Reasoning):
    \begin{enumerate}
        \item First Hop:
        \item Second Hop (Justify how it relies on previous hops and by which reasoning type):
        \ldots
    \end{enumerate}
    \item Justification why the question is Multimodal
    \item Justification why the question is Multi-Episode:
    \item List of Episode involved (separated by comma). Example 6 January 2003, ...:
    \item Answer to the above question in 200 words maximum:
    \item List of Characters involved in the question (separated by comma):
\end{enumerate}

\end{tcolorbox}

\subsubsection{Example Output}

\begin{tcolorbox}[breakable, colback=white, colframe=gray!20, left=2pt,  coltitle=black,  title=\textbf{}]

\begin{enumerate}
    \item Multimodal Multi-Episode Multi-Hop Question: Considering Carly's initial excitement about the "Midnight Sun" vodka venture with Craig, how did the visual presence of the product samples at her home lead to a dangerous situation for her children, and what was her subsequent emotional realization and decision regarding the business, particularly in light of her past experiences with an alcoholic father, as evidenced by the blackened kitchen appliances?\\
    \item Compact Question: How did "Midnight Sun" vodka samples at Carly's home lead to danger for her children, and what was her decision about the business, given her past with an alcoholic father and the visual evidence of the incident?\\
    \item Justification why the question is Multi-Hop:\\
        \begin{enumerate}
            \item First Hop (Bridge Entities): Identify Carly's involvement in the "Midnight Sun" vodka business with Craig. This establishes the core entity and her initial engagement.
            \item Second Hop (Causal \& Counterfactual): Connect the vodka samples (a specific product of the business) being present at her home to her children, Parker and Liberty, consuming them and getting drunk. This links the business to a direct consequence involving her family.
            \item Third Hop (Causal \& Counterfactual / Temporal Reasoning): Link the children's consumption and subsequent actions (leaving pizza in the oven) to the fire and the visually evident "blackened kitchen appliances and counters," which then triggers Carly's emotional realization of the danger.
            \item Fourth Hop (Causal \& Counterfactual): Explain Carly's decision regarding the "Midnight Sun" business, specifically in light of her past trauma with an alcoholic father and the recent dangerous incident involving her children.
        \end{enumerate}

    \item Justification why the question is Multimodal:
    The question explicitly refers to "visual presence of the product samples" (the vodka cans/bottles) and "blackened kitchen appliances and counters" as evidence of the incident and its impact on Carly. The act of "drinking" is also an action that would be visually depicted.

    \item Justification why the question is Multi-Episode:
    The events described span multiple episodes: the inception of the "Midnight Sun" vodka business (09 February), the children consuming the samples and Carly's initial realization (20 February), and the subsequent fire and Carly's renewed emotional distress upon seeing the damaged kitchen (04 March).

    \item List of Episode involved: 09 February 2009, 20 February 2009, 04 March 2009

    \item Answer to the above question in 200 words maximum:
Carly, initially excited about her "Midnight Sun" vodka venture with Craig, found herself in a dangerous situation when Craig brought unlabeled product samples to her home. Her children, Parker and Liberty, mistook these for soda, drank them, and became drunk. This led to them ditching school and leaving a pizza in the oven, causing a fire and leaving her kitchen with "blackened appliances and counters."

Upon seeing the damage and realizing she could have lost her son, Carly was deeply shocked and emotional. She revealed her childhood trauma of being terrorized by an alcoholic father, expressing her fear that Parker might follow a similar path. Despite this profound realization and the incident's severity, Carly initially decided not to shelve the "Midnight Sun" project, affirming her partnership with Craig. However, the event profoundly underscored her deep-seated fears about alcohol's impact on her family.

    \item List of Characters involved in the question: Carly Tenney, Craig Montgomery, Parker Snyder, Liberty Snyder, Johnny Donovan, Jack Snyder, Janet Snyder
\end{enumerate}

\end{tcolorbox}

\subsection{Multi-Hop Filtering} \label{app:multi_hop_filtering}

We provide the prompt below to filter our generated QA pairs, keeping only the questions that are truly multi-hop. The prompt is given to Qwen3-30B-thinking.

\begin{itemize}
\item {\color{red} <QUESTION>} is the question to classify as multi-hop or not.
\end{itemize}

\begin{tcolorbox}[colback=white, colframe=gray!20, left=2pt,  coltitle=black,  title=\textbf{}]

Justify whether this question is truly multi-hop. you assume you have access to a document containing all the answers. No external knowledge is required.\\

Definition of Multi-Hop:

\begin{enumerate}
    \item A question is multi-hop if it requires reasoning across at least two pieces of information to form a complete answer.
    \item Requirements
    \begin{enumerate}
        \item Inference required: Answers cannot be copied verbatim; they demand logical connection.
        \item The Hops can be part of the same narrative
    \end{enumerate}
\end{enumerate}

You can compare the pros and cons and provide your final justification and answer.

Question: {\color{red} <QUESTION>}

\end{tcolorbox}

\subsection{Video Planners Prompts} \label{app:baseline_prompting}

We provide the prompts below for the video planners we tested on~\textsc{SagaQA}. The prompts are given to either Qwen3-30B-instruct or Mistral-Small-3.1-24B.

\subsubsection{Parallel Planner~(Prompt~1)}

\begin{itemize}
\item {\color{red} <QUESTION>} is a composite question about the TV show.
\end{itemize}

\begin{enumerate}[label=\alph*), start=1]
    \item \textbf{Keywords Extraction}\\We start by extracting all the keywords from the question to answer about the TV show.
\end{enumerate}

\begin{tcolorbox}[colback=white, colframe=gray!20, left=2pt,  coltitle=black,  title=\textbf{}]
Question: {\color{red} <QUESTION>}\\

Decompose the question into keywords to the video. Keep only the most relevant keywords. You may drop non useful keywords that does not add to the meaning.\\

Keywords List:\\
1. ...\\
2. ...\\
...
\end{tcolorbox}

\begin{enumerate}[label=\alph*), start=2]
    \item \textbf{Keywords Ordering}\\Given the list of keywords we previously extracted, we can reorganize them by difficulty.
\end{enumerate}

\begin{tcolorbox}[colback=white, colframe=gray!20, left=2pt,  coltitle=black,  title=\textbf{}]

Order the keywords from the easiest to match to a unique scene or video moment to the least easy to match:\\

Ordered Keywords List:\\
1. ...\\
2. ...\\
...
\end{tcolorbox}

\begin{enumerate}[label=\alph*), start=3]
    \item \textbf{Keywords Grouping}\\We group the keywords that are related together.
\end{enumerate}

\begin{tcolorbox}[colback=white, colframe=gray!20, left=2pt,  coltitle=black,  title=\textbf{}]
Group the keywords together. Do not miss any keyword.\\

Grouped keywords:\\
** Group 1: ... **\\
1. ...\\
2. ...\\
...\\
    
** Group 2: ... **\\
1. ...\\
2. ...\\
...
\end{tcolorbox}

\begin{enumerate}[label=\alph*), start=4]
    \item \textbf{Subquestion Generation for each Group}\\We generate a title for each group of keywords. This title is later used as a query to the VideoRAG.
\end{enumerate}

\begin{tcolorbox}[colback=white, colframe=gray!20, left=2pt,  coltitle=black,  title=\textbf{}]
Propose a title for each group in a few words. Always include the group keywords into the title. Always refer to the name of the characters in your title.\\

Group List:\\

** Group 1 Title: ... **\\
** Group 2 Title: ... **\\
...
\end{tcolorbox}

\subsubsection{Parallel Planner~(Prompt~2)} \label{app:prompt_parallel_planner}

\begin{itemize}
\item {\color{red} <QUESTION>} is a composite question about the TV show.
\end{itemize}

\begin{tcolorbox}[breakable, colback=white, colframe=gray!20, left=2pt,  coltitle=black,  title=\textbf{}]
Task Definition for the Querying Agent (Planner)\\

You are a Video Query Planner Agent. Your role is to retrieve information from a video in order to answer a complex, multi-hop question.\\
Your task is not to answer the question directly. Instead, you must plan and issue a sequence queries to the video.\\

Objective\\

Given a Question, you must decompose it into simple video queries that, when answered sequentially, will provide all necessary evidence to fully answer the question.\\

Querying Rules\\

\begin{enumerate}
    \item A query is always grounded into the question. Do not make up any new information not already present within the question.
    \item All queries should target a disctinct information (non-overlapping).
    \item Regroup similar queries together. Do not over decompose the queries in too specific.
    \item Queries should be very short and clear.
    \item Rank your queries from easier to harder
\end{enumerate}

Output Format\\

Do not include answers, summaries, or explanations — only the planned queries.\\
Your output must be in the same format as in the examples below\\

Example:\\

Question: What visual actions directly resulted in the definitive exposure of Meg's deception about taking her medication, and who was the primary witness to this final revelation?\\

Queries:\\
* Query 1: Meg's deception about taking her medication.\\
* Query 2: Visual actions that exposed Meg's deception about taking her medication.\\
* Query 3: The primary witness to the revelation of Meg's deception about taking her medication.\\

Question:\\
{\color{red} <QUESTION>}

Queries:\\
* Query 1: ...\\
* Query 2: ...\\
...
\end{tcolorbox}

\subsubsection{Sequential Planner} \label{app:prompt_sequential_planner}

At any given step~$i$, we proceed as follows:
\begin{itemize}
    \item Generate the subquestion for the~$i^{\text{th}}$ iteration.
    \item Answer subquestion~$i$ by querying the VideoRAG.
\end{itemize}

Note that we do not need to produce a subquestion for the first iteration, since the first subquestion is simply the main question.

Below, we provide the prompt given to the sequential planner agent at step~$i$. As stated in Section~\ref{subsec:implementation_details}, we limit the maximum depth of the reasoning sequence to~15 as we observe that reasoning beyond this limit does not yield further improvements.

\begin{itemize}
    \item {\color{red} <QUESTION>} is a composite question about the TV show.
    \item {\color{red} <SUBQUESTION\_i>} is the subquestion generated by the sequential planner at step~$i$.
    \item {\color{red} <ANSWER\_i>} is the answer produced by the VideoRAG to the~$i^{\text{th}}$ subquestion.
\end{itemize}

\begin{tcolorbox}[breakable, colback=white, colframe=gray!20, left=2pt,  coltitle=black,  title=\textbf{}]
Query 1: {\color{red} <QUESTION>}\\
Answer to Query 1: {\color{red} <ANSWER\_1>}\\

Query 2: {\color{red} <SUBQUESTION\_2>}\\
Answer to Query 2: {\color{red} <ANSWER\_2>}\\

\ldots

Query i - 1: {\color{red} <SUBQUESTION\_i - 1>}\\
Answer to Query i - 1: {\color{red} <ANSWER\_i - 1>}\\

You are a Video Query Planner Agent. Your goal is to answer Query 1.\\
Your task is not to answer Query 1 directly. Instead, you must plan and issue a sequence of queries to eventually answer Query 1.\\

Given the answers to the previous queries, you can choose between different actions:\\

\begin{enumerate}
    \item BREAK DOWN the last query (e.g. I need to break down the last query into subqueries as the last query is too complex).
    \item MOVE\_ON to the next query (e.g. the answer to the last query is satisfying and we can move on to the next query or we decide to move on after too many unsuccessful attempts to answer the last query).
\end{enumerate}

Justify which action you want to perform. Given your action, what is the next query. Produce the next query only.\\

Querying Rules\\

\begin{enumerate}
    \item Altogether, the queries should eventually help you to build your answer to Query 1.
    \item A query is always grounded into the question. Do not make up any new information not already present within the question.
    \item The Queries should be very short and clear.
    \item Multiple BREAK DOWNS can be performed as long as they are justified.
\end{enumerate}

Answer in the following way:\\

Justification: ...\\
Action performed: ...\\
Query i: ...
\end{tcolorbox}

\subsubsection{Hybrid Planner}

The hybrid planner is a combination of both the parallel and sequential planners. We leverage the prompts of both methods in the implementation of the hybrid approach.

At any given step~$i$, we proceed as follows:
\begin{itemize}
    \item Prompt the agent for the next action~(\texttt{BREAKDOWN} or~\texttt{MOVE\_ON}), using the same prompt as the sequential planner in~Appendix~\ref{app:prompt_sequential_planner}
    \item If the action is~\texttt{BREAKDOWN}, we use the prompt from the parallel planner to decompose the current query into a list of independent subquestions~(Appendix~\ref{app:prompt_parallel_planner}).
    \item If the action is \texttt{MOVE\_ON}, the agent processes the next available subquestion from the most recent \texttt{BREAKDOWN}. If no pending subquestions remain across all prior \texttt{BREAKDOWN} steps, the reasoning process stops.
\end{itemize}

As specified in Section~\ref{subsec:implementation_details}, we limit the maximum depth of the reasoning tree to~2 and do not allow any reasoning beyond that fixed limit.

\subsubsection{Answer Synthesis Prompt} \label{app:answer_synthesis_prompt}

We provide below the prompt used by the video planners to generate their final answers. We ask the LLM to produce an answer in~200 words.

\begin{itemize}
    \item {\color{red} <QUESTION>} is a composite question about the TV show.
    \item {\color{red} <SUBQUESTION\_i>} is the subquestion generated by the video planner at the~$i^{\text{th}}$ step.
    \item {\color{red} <ANSWER\_i>} is the answer produced by the VideoRAG to the~$i^{\text{th}}$ subquestion.
\end{itemize}

\begin{tcolorbox}[breakable, colback=white, colframe=gray!20, left=2pt,  coltitle=black,  title=\textbf{}]
Query: {\color{red} <SUBQUESTION\_1>}

Answer in 200 words maximum: {\color{red} <ANSWER\_1>}

Query: {\color{red} <SUBQUESTION\_2>}

Answer in 200 words maximum: {\color{red} <ANSWER\_2>}

\ldots

Query: {\color{red} <QUESTION>}

Answer in 200 words maximum: ...
\end{tcolorbox}

\section{Additional Experiments} \label{app:experiments}

In Table~\ref{tab:main-results-MFactSum-1.5}, we report the results using Mistral-Small-3.1-24B as both the planner and question-answering models. Our conclusions remain the same as in Section~\ref{subsec:clip_results} where we used Qwen3-30B-Instruct~(see Table~\ref{tab:main-results-MFactSum})

\begin{table*}
\center
\setlength{\tabcolsep}{2pt}
\resizebox{0.90\textwidth}{!}{
 \begin{tabular}{@{}lcccc|ccc@{}}
\toprule
& r1 & r2 & rlsum & METEOR & Precision & Recall & F1 \\
\midrule
\multicolumn{1}{c}{\textbf{Non-Planning}} & & & & & & & \\
\multicolumn{1}{l}{TextRAG} & 36.02 & 6.91 & 32.87 & 24.02 & 29.17 & 11.14 & 15.06 \\
\multicolumn{1}{l}{VideoRAG} &~\textbf{37.64} &~\textbf{7.56} &~\textbf{34.19} & 24.05 & 37.95 & 30.04 & 28.30 \\
\noalign{\vskip 0.5ex}
\cdashline{1-8}[2pt/2pt]
\noalign{\vskip 0.5ex}
\multicolumn{1}{c}{\textbf{Parallel Planners}} & & & & & & & \\
\multicolumn{1}{l}{\textit{Parallel Planner (Prompt 1) + VideoRAG}} & 36.74 & 7.41 & 33.29 &~\textbf{25.04} & 38.11 & 26.54 & 27.39 \\
\multicolumn{1}{l}{\textit{Parallel Planner (Prompt 2) + VideoRAG}} & 36.51 & 7.34 & 33.06 & 25.00 &~\textbf{56.62} & 34.14 & 37.74 \\
\noalign{\vskip 0.5ex}
\cdashline{1-8}[2pt/2pt]
\noalign{\vskip 0.5ex}
\multicolumn{1}{c}{\textbf{Sequential Planners}} & & & & & & & \\
\multicolumn{1}{l}{\textit{Sequential Planner + VideoRAG}} & 34.70 & 7.26 & 31.54 & 23.00 & 48.82 & 31.87 & 34.21 \\
\noalign{\vskip 0.5ex}
\cdashline{1-8}[2pt/2pt]
\noalign{\vskip 0.5ex}
\multicolumn{1}{c}{\textbf{Hybrid Planners}} & & & & & & & \\
\multicolumn{1}{l}{\textit{Hybrid Planner + VideoRAG}} & 34.14 & 6.80 & 30.98 & 24.01 & 48.80 &~\textbf{45.70} &~\textbf{41.69} \\

\bottomrule
\end{tabular}
}
\caption{\textbf{Results on the test set with Mistral-Small-3.1-24B.} Column descriptions are the same as in Table~\ref{tab:main-results-MFactSum}. Best results are in \textbf{bold}.}
\label{tab:main-results-MFactSum-1.5}
\end{table*}

\section{Reasoning Trace Examples} \label{app:reasoning_trace}

We compare the reasoning traces of the hybrid planner with those of two sequential planners on a question sampled from our dataset.

\begin{tcolorbox}[breakable, colback=white, colframe=gray!20, left=2pt,  coltitle=black,  title=\textbf{}]
\textbf{\underline{Composite Question:} After Alison confronted Mick in the belfry about his actions towards Dr. Cantor, what visual action did Mick perform that indicated his guilt, and how did this specific visual action later manifest in Alison's disturbed dream on January 13th?}
\end{tcolorbox}

The reasoning traces of the two sequential methods below illustrate how sequential planning can easily lead to repetitive queries, causing the agent to revisit similar queries instead of effectively covering all aspects of a multi-hop question. We use the same color to underlight semantically similar queries, illustrating the agent’s failure to sufficiently explore new queries.

\subsubsection{VideoExplorer (Sequential Planner)}

\begin{tcolorbox}[breakable, colback=white, colframe=gray!20, left=2pt,  coltitle=black,  title=\textbf{}]
    \textbf{Q1:} \coluline{red}{Locate the segment where Alison confronts Mick in the belfry about his actions towards Dr. Cantor. After this confrontation, what visual action did Mick perform that indicated his guilt?}\\

    \textbf{Q2:} \coluline{red}{Locate the segment where Mick is in the belfry and related to Dr. Cantor. After this segment, what visual action did Mick perform that indicated his guilt?}\\

    \textbf{Q3:} \coluline{red}{Locate the segment where Mick is in the belfry. After this segment, what visual action did Mick perform that indicated his guilt?}\\

    \textbf{Q4:} \coluline{blue}{Locate the segment where Alison has a dream on January 13th. Describe the content of the dream, especially any visual actions by Mick that might indicate guilt.}\\

    \textbf{Q5:} Locate the segment where Mick appears. Describe his actions or context.\\

    \textbf{Q6:} \coluline{green}{Locate the segment where Alison appears. Describe her actions or context.}\\

    \textbf{Q7:} \coluline{blue}{Locate segments depicting dreams, nightmares, or similar sleep-related scenarios, particularly in a belfry setting. Describe the visual actions of characters within these dreams or nightmares, if any characters resemble Mick, and any visual actions that might indicate guilt.}\\

    \textbf{Q8:} \coluline{green}{Locate the segment where Alison appears. Describe her actions or context.}\\

    \textbf{Q9:} \coluline{blue}{Locate the segment where Alison confronts Mick in a forested area. After this confrontation, what visual action does Mick perform that indicates his guilt?}\\

    \textbf{Q10:} Is Alison confronting Mick in this segment? If so, what visual action does Mick perform that indicates his guilt?\\
\end{tcolorbox}

\subsubsection{Sequential Planner + VideoRAG}

\begin{tcolorbox}[breakable, colback=white, colframe=gray!20, left=2pt,  coltitle=black,  title=\textbf{}]
\textbf{Q1:} The visual action Mick performed after Alison confronted him in the belfry about his actions towards Dr. Cantor, and how that action manifested in Alison's disturbed dream on January 13th.\\
ACTION: BREAK DOWN\\

\textbf{Q2:} \coluline{red}{What specific visual action did Mick perform in the belfry after Alison confronted him about Dr. Cantor?}\\
ACTION: BREAK DOWN\\

\textbf{Q3:} \coluline{red}{What specific gesture or movement did Mick make immediately after Alison confronted him in the belfry?}\\
ACTION: BREAK DOWN\\

\textbf{Q4:} \coluline{red}{What visual detail or action occurred in the belfry immediately after the confrontation between Alison and Mick?}\\
ACTION: BREAK DOWN\\

\textbf{Q5:} \coluline{red}{What specific visual element in the belfry (e.g., bell, stone arch, shadow) was present during the confrontation between Alison and Mick?}\\
ACTION: BREAK DOWN\\

\textbf{Q6:} What distinctive background feature (e.g., large bell, cracked stone wall, hanging rope) was visible in the belfry during the confrontation?\\
ACTION: BREAK DOWN\\

\textbf{Q7:} What camera angle or movement (e.g., close-up, slow pan, overhead shot) was used during the scene in the belfry?\\
ACTION: MOVE ON\\

\textbf{Q8:} What specific visual element appeared in Alison’s dream on January 13th that could be linked to Mick’s actions in the belfry?\\
ACTION: BREAK DOWN\\

\textbf{Q9:} \coluline{blue}{What object or image repeatedly appeared in Alison’s dream on January 13th?}\\
ACTION: BREAK DOWN\\

\textbf{Q10:} \coluline{blue}{What single recurring object (e.g., bell, rope, shadow, hand gesture) appeared in Alison’s dream on January 13th?}\\
ACTION: MOVE ON\\

\textbf{Q11:} What specific physical gesture or movement did Mick make in the belfry that could indicate guilt (e.g., covering his face, looking away, trembling hand)?\\
ACTION: BREAK DOWN\\

\textbf{Q12:} \coluline{green}{Did Mick touch or interact with any object in the belfry immediately after the confrontation?}\\
ACTION: BREAK DOWN\\

\textbf{Q13:} \coluline{green}{Did Mick touch the bell in the belfry immediately after the confrontation with Alison?}
\\

\end{tcolorbox}

\subsubsection{Hybrid Planner}

\begin{tcolorbox}[breakable, colback=white, colframe=gray!20, left=2pt,  coltitle=black,  title=\textbf{}]

\textbf{Q1:} The events involving Alison confronting Mick in the belfry about his actions towards Dr. Cantor.\\
ACTION: BREAKDOWN\\
\textbf{Q1':} The events that took place in the belfry, specifically involving Alison and Mick.\\
ACTION: MOVE ON\\
\textbf{Q2':} Mick's actions towards Dr. Cantor.\\
ACTION: MOVE ON\\
\textbf{Q3':} The location of the confrontation between Alison and Mick.\\
ACTION: MOVE ON\\

\textbf{Q2:} Alison's reasons for confronting Mick about his actions towards Dr. Cantor.\\
ACTION: BREAKDOWN\\
\textbf{Q1'':} \coluline{red}{The visual action Mick performed in the belfry that indicated his guilt.}\\
ACTION: MOVE ON\\
\textbf{Q2'':} \coluline{blue}{Mick's actions in the belfry.}\\
ACTION: MOVE ON\\
\textbf{Q3'':} \coluline{red}{Visual actions performed by Mick in the belfry that suggest guilt.}\\
ACTION: BREAKDOWN\\

\textbf{Q3:} \coluline{green}{How the visual action Mick performed in the belfry manifested in Alison's dream on January 13th.}\\
ACTION: MOVE ON\\
\textbf{Q1''':} \coluline{blue}{Mick's visual actions in the belfry.}\\
ACTION: MOVE ON\\
\textbf{Q2''':} Alison's dream on January 13th.\\
ACTION: MOVE ON\\
\textbf{Q3''':} \coluline{green}{How Mick's visual actions in the belfry manifested in Alison's dream on January 13th.}\\

\end{tcolorbox}

\end{document}